%% file: arxiv.tex
\documentclass[final]{cvpr}

\usepackage{times}
\usepackage{epsfig}
\usepackage{graphicx}
\usepackage{amsmath}
\usepackage{amssymb}

\usepackage{url}
\usepackage{cite}
\usepackage{multirow}
\usepackage{booktabs}
\usepackage{tabulary}
\usepackage[dvipsnames]{xcolor}
\usepackage[caption=false]{subfig}
\usepackage[british,english,american]{babel}
\usepackage{soul}
\usepackage{xspace}\xspace
\usepackage{adjustbox}
\usepackage{array}

\usepackage{dblfloatfix}
\usepackage{transparent}

\usepackage{algorithm}
\usepackage{listings}
\usepackage{pifont}

\usepackage[pagebackref=true,breaklinks=true,colorlinks]{hyperref}

\newcommand{\model}{SETR}
\newcommand{\modelFull}{\textit{SEgmentation TRansformer}}

\newcommand{\naiveModel}{SETR-\textit{Na\"ive}}
\newcommand{\pupModel}{SETR-\textit{PUP}}
\newcommand{\mlaModel}{SETR-\textit{MLA}}

\newcommand{\smallNaiveModel}{SETR-\textit{Na\"ive-Base}}
\newcommand{\smallMlaModel}{SETR-\textit{MLA-Base}}
\newcommand{\smallPupModel}{SETR-\textit{PUP-Base}}

\newcommand{\NaiveDeit}{SETR-\textit{Na\"ive-DeiT}}
\newcommand{\MlaDeit}{SETR-\textit{MLA-DeiT}}
\newcommand{\PupDeit}{SETR-\textit{PUP-DeiT}}

\newcommand{\hybridModelB}{\textit{Hybrid-Base}}
\newcommand{\hybridModelD}{\textit{Hybrid-DeiT}}
\newcommand{\hybridModel}{\textit{Hybrid}}

\def\eg{\textit{e.g.}}
\def\ie{\textit{i.e.}}

\makeatletter\renewcommand\paragraph{\@startsection{paragraph}{4}{\z@}
  {.5em \@plus1ex \@minus.2ex}{-.5em}{\normalfont\normalsize\bfseries}}\makeatother

\newcolumntype{x}[1]{>{\centering\arraybackslash}p{#1pt}}
\newcolumntype{y}[1]{>{\raggedright\arraybackslash}p{#1pt}}
\newcolumntype{z}[1]{>{\raggedleft\arraybackslash}p{#1pt}}
\newlength\savewidth\newcommand\shline{\noalign{\global\savewidth\arrayrulewidth
  \global\arrayrulewidth 1pt}\hline\noalign{\global\arrayrulewidth\savewidth}}
\newcommand{\tablestyle}[2]{\setlength{\tabcolsep}{#1}\renewcommand{\arraystretch}{#2}\centering\footnotesize}

\begin{document}

\title{Rethinking Semantic Segmentation from a Sequence-to-Sequence Perspective with Transformers}

\author{Sixiao Zheng\textsuperscript{1}\thanks{Work done while Sixiao Zheng was interning at Tencent Youtu Lab.}
\quad
Jiachen Lu\textsuperscript{1}
\quad
Hengshuang Zhao\textsuperscript{2} 
\quad
Xiatian Zhu\textsuperscript{3} 
\quad
Zekun Luo\textsuperscript{4}
\quad
Yabiao Wang\textsuperscript{4} \\
\quad
Yanwei Fu\textsuperscript{1} 
\quad
Jianfeng Feng\textsuperscript{1} 
\quad
Tao Xiang\textsuperscript{3,~5} 
\quad
Philip H.S. Torr\textsuperscript{2} 
\quad
Li Zhang\textsuperscript{1}\thanks{Li Zhang (lizhangfd@fudan.edu.cn) is the corresponding author with School of Data Science, Fudan University.} \\
\textsuperscript{1}Fudan University
\quad
\textsuperscript{2}University of Oxford
\quad
\textsuperscript{3}University of Surrey \\
\quad
\textsuperscript{4}Tencent Youtu Lab
\quad
\textsuperscript{5}Facebook AI\\
\url{https://fudan-zvg.github.io/SETR}
}

\maketitle

\input{files/0-abstract}

\input{files/1-intro}

\input{files/2-related_work}

\input{files/3-method}

\input{files/4-experiment}

\input{files/5-conclusion}

{\small
\bibliographystyle{ieee_fullname}
\bibliography{egbib}
}

\clearpage

\appendix
\section*{Appendix}

\input{files/6-appendix}

\end{document}

%% file: files/0-abstract.tex

\begin{abstract}
Most recent semantic segmentation methods adopt a fully-convolutional network (FCN) with an encoder-decoder architecture. The encoder progressively reduces the spatial resolution and learns more abstract/semantic visual concepts with larger receptive fields.  Since context modeling is critical for segmentation, the latest efforts have been focused on increasing the receptive field, through either dilated/atrous convolutions or inserting attention modules. However, the encoder-decoder based FCN architecture remains unchanged. In this paper, we aim to provide an alternative perspective by treating semantic segmentation as a sequence-to-sequence prediction task. Specifically, we deploy a pure transformer (\ie, without convolution and resolution reduction) to encode an image as a sequence of patches.  With the global context modeled in every layer of the transformer, this encoder can be combined with a simple decoder to provide a powerful segmentation model, termed SEgmentation TRansformer (SETR). 
Extensive experiments show that SETR achieves new state of the art on ADE20K (50.28\% mIoU), Pascal Context (55.83\% mIoU) and competitive results on Cityscapes.
Particularly, we achieve the {\em first} position in the highly competitive ADE20K test server 
leaderboard on the day of submission.
\end{abstract}

%% file: files/1-intro.tex
\section{Introduction}

Since the seminal work of \cite{fcn}, existing semantic segmentation models have been dominated by those based on fully convolutional network (FCN). 
A standard FCN segmentation model has an encoder-decoder architecture:
the {\em encoder} is for feature representation learning, while
the {\em decoder} for pixel-level classification of the feature representations
yielded by the encoder.
Among the two,  feature representation learning (\ie, the encoder)
is arguably the most important model component \cite{chen2018deeplab,pspnet,zhang2020dynamic,li2020improving}.
The encoder, like most other CNNs designed for image understanding, consists of stacked convolution layers. Due to concerns on computational cost, the resolution of feature maps is reduced progressively, and the encoder is hence able to learn more abstract/semantic visual concepts with a gradually increased receptive field. Such a design is popular due to  
two favorable merits,
namely translation equivariance and locality.
The former respects well the nature of imaging process
\cite{zhang2019making} which underpins the model generalization
ability to unseen image data.
Whereas the latter controls the model complexity 
by sharing parameters across space. However, 
it also raises a fundamental limitation
that learning {long-range dependency information}, critical for semantic segmentation in unconstrained scene images \cite{segnet,denseaspp},
becomes challenging due to still limited receptive fields.

To overcome this aforementioned limitation,  a number of approaches have been introduced recently.
One approach is to directly manipulate the convolution operation. This includes
large kernel sizes \cite{largekernel}, 
atrous convolutions \cite{holschneider1990real,chen2018deeplab},
and image/feature pyramids \cite{pspnet}.
The other approach is to integrate attention modules into the FCN architecture. Such a module aims to model {\em global} interactions of all pixels in the feature map \cite{wang2018nonlocal}.
When applied to 
semantic segmentation \cite{huang2018ccnet,li2019global}, a common design is to combine the attention module to the FCN architecture with attention layers sitting on the top.
Taking either approach, the standard encoder-decoder FCN model architecture remains unchanged. More recently, attempts have been made to get rid of convolutions altogether and deploy attention-alone models \cite{wang2020axial} instead. However, even without convolution, they do not change the nature of the FCN model structure: an encoder downsamples the spatial resolution of the input, developing lower-resolution feature mappings useful for discriminating semantic classes, and the decoder upsamples the feature representations into a full-resolution segmentation map.

In this paper, we aim to provide a rethinking to the semantic segmentation model design and contribute an alternative. In particular, we propose to replace the stacked convolution layers based encoder with gradually reduced spatial resolution with a pure transformer \cite{vaswani2017attention}, resulting in a new segmentation model termed {\em SEgmentation TRansformer} (SETR). This transformer-alone encoder treats an input image as a sequence of {\em image patches} represented by learned patch embedding, and transforms the sequence with global self-attention modeling for discriminative feature representation learning. Concretely, we first decompose an image into a grid
of fixed-sized patches, forming a sequence of patches.
With a linear embedding layer applied to the 
flattened pixel vectors of every patch, we then obtain a 
sequence of feature embedding vectors
as the input to a transformer. Given the learned features from the encoder transformer, 
a decoder is then used 
to recover the original image resolution. 
Crucially there is {\em no} downsampling in spatial resolution but global context modeling at every layer of the encoder transformer, thus offering a completely new perspective to the semantic segmentation problem.   

This pure transformer design is inspired by its tremendous success in natural language processing (NLP)~\cite{vaswani2017attention,Devlin2018}. 
More recently, a pure vision transformer or ViT \cite{dosovitskiy2020image} has shown to be effective for image classification tasks.  It thus provides direct evidence that the traditional stacked convolution layer (\ie, CNN) design can be challenged and image features do not necessarily need to be learned progressively from local to global context by reducing spatial resolution. However, extending a pure transformer 
from image classification to a spatial location sensitive task of semantic segmentation is non-trivial. We show empirically that SETR not only offers a new perspective in model design, but also achieves new state of the art on a number of benchmarks.

The following {\bf contributions} are made in this paper:
(1) We reformulate the image semantic segmentation problem
from a {\em sequence-to-sequence} learning perspective, offering an alternative to the dominating encoder-decoder FCN model design. 
(2) As an instantiation, we exploit the transformer framework
to implement our fully attentive feature representation encoder
by sequentializing images.
(3) To extensively examine the self-attentive feature presentations,
we further introduce three different decoder designs
with varying complexities.
Extensive experiments show that our SETR models can 
learn superior feature representations as compared to 
different FCNs with and without attention modules, yielding 
new state of the art on ADE20K (50.28\%), Pascal Context (55.83\%) and competitive results on Cityscapes.
Particularly, our entry is ranked the $1^{st}$ place in the highly competitive ADE20K test server leaderboard.

%% file: files/2-related_work.tex
\section{Related work}

\input{figure/sketch_transformer}

\paragraph{Semantic segmentation}
Semantic image segmentation has been significantly boosted with the development of deep neural networks. By removing  fully connected layers, the fully convolutional network (FCN)~\cite{fcn} is able to achieve pixel-wise predictions. While the predictions of FCN are relatively coarse, several CRF/MRF~\cite{chen2015semantic,liu2015semantic,zheng2015conditional} based approaches are developed to help refine the coarse predictions. To address the inherent tension between semantics and location~\cite{fcn}, coarse and fine layers need to be aggregated for both the encoder and decoder. This leads to different variants of the encoder-decoder structures~\cite{unet,deconvnet,segnet} for multi-level feature fusion.

Many recent  efforts have been focused on addressing the limited receptive field/context modeling problem in FCN.  To enlarge the receptive field, DeepLab~\cite{deeplabv1} and Dilation~\cite{dilation} introduce the dilated convolution. Alternatively, context modeling is the focus of PSPNet~\cite{pspnet} and DeepLabV2~\cite{deeplabv2}. The former proposes the PPM module to obtain different region's contextual information while the latter develops ASPP module that adopts pyramid dilated convolutions with different dilation rates. Decomposed large kernels~\cite{largekernel} are also utilized for context capturing.
Recently, attention based models are popular for capturing long range context information. PSANet~\cite{psanet} develops the pointwise spatial attention module for dynamically capturing the long range context. DANet~\cite{DAnet} embeds both spatial attention and channel attention. CCNet~\cite{huang2019ccnet} alternatively focuses on economizing the heavy computation budget introduced by full spatial attention. DGMN~\cite{zhang2020dynamic} builds a dynamic graph message passing network for scene modeling and it can significantly reduce the computational complexity. Note that all these approaches are still based on FCNs where the feature encoding and extraction part are based on classical ConvNets like VGG~\cite{vgg} and ResNet~\cite{resnet}. In this work, we alternatively rethink the semantic segmentation task from a different perspective.

\paragraph{Transformer}
Transformer and self-attention models have revolutionized  machine translation and NLP~\cite{vaswani2017attention,Devlin2018,dai2019transformer,Yang2019xlnet}. Recently, there are also some explorations for the usage of transformer structures in image recognition. Non-local network~\cite{wang2018nonlocal} appends transformer style attention onto the convolutional backbone. AANet~\cite{bello2019attention} mixes convolution and self-attention for backbone training. LRNet~\cite{hu2019local} and stand-alone networks~\cite{ramachandran2019stand} explore local self-attention to avoid the heavy computation brought by global self-attention. SAN~\cite{zhao2020san} explores two types of self-attention modules. Axial-Attention~\cite{wang2020axial} decomposes the global spatial attention into two separate axial attentions such that the computation is largely reduced. Apart from these pure transformer based models, there are also CNN-transformer hybrid ones. DETR~\cite{carion2020end} and the following deformable version utilize transformer for object detection where transformer is appended inside the detection head. STTR~\cite{li2020revisiting} and LSTR~\cite{liu2020lane} adopt transformer for disparity estimation and lane shape prediction respectively.  Most recently, ViT~\cite{dosovitskiy2020image} is the first work to show that a pure transformer based image classification model can achieve the state-of-the-art. It provides direct inspiration to exploit a pure transformer based encoder design in a semantic segmentation model.

The most related work is \cite{wang2020axial} which  also leverages attention for image segmentation. However, there are several key differences. First, though convolution is completely removed in \cite{wang2020axial} as in our SETR, their model still follows the conventional FCN design in that spatial resolution of feature maps is reduced progressively. In contrast, our sequence-to-sequence prediction model keeps the same spatial resolution throughout and thus represents a step-change in model design. 
Second, to maximize the scalability on modern hardware accelerators and facilitate easy-to-use, we stick to the standard self-attention design. Instead, \cite{wang2020axial} adopts a specially designed axial-attention  \cite{ho2019axial} which is less scalable to standard computing facilities. Our model is also superior in segmentation accuracy (see Section~\ref{sec:exp}). 

%% file: figure/sketch_transformer.tex
\begin{figure*}[t]
\begin{minipage}[b]{.45\linewidth}
    \centering
    \subfloat[][]{\label{Genelecs:Genelec 8010 AP}\includegraphics[height=7cm, width=0.9\linewidth]{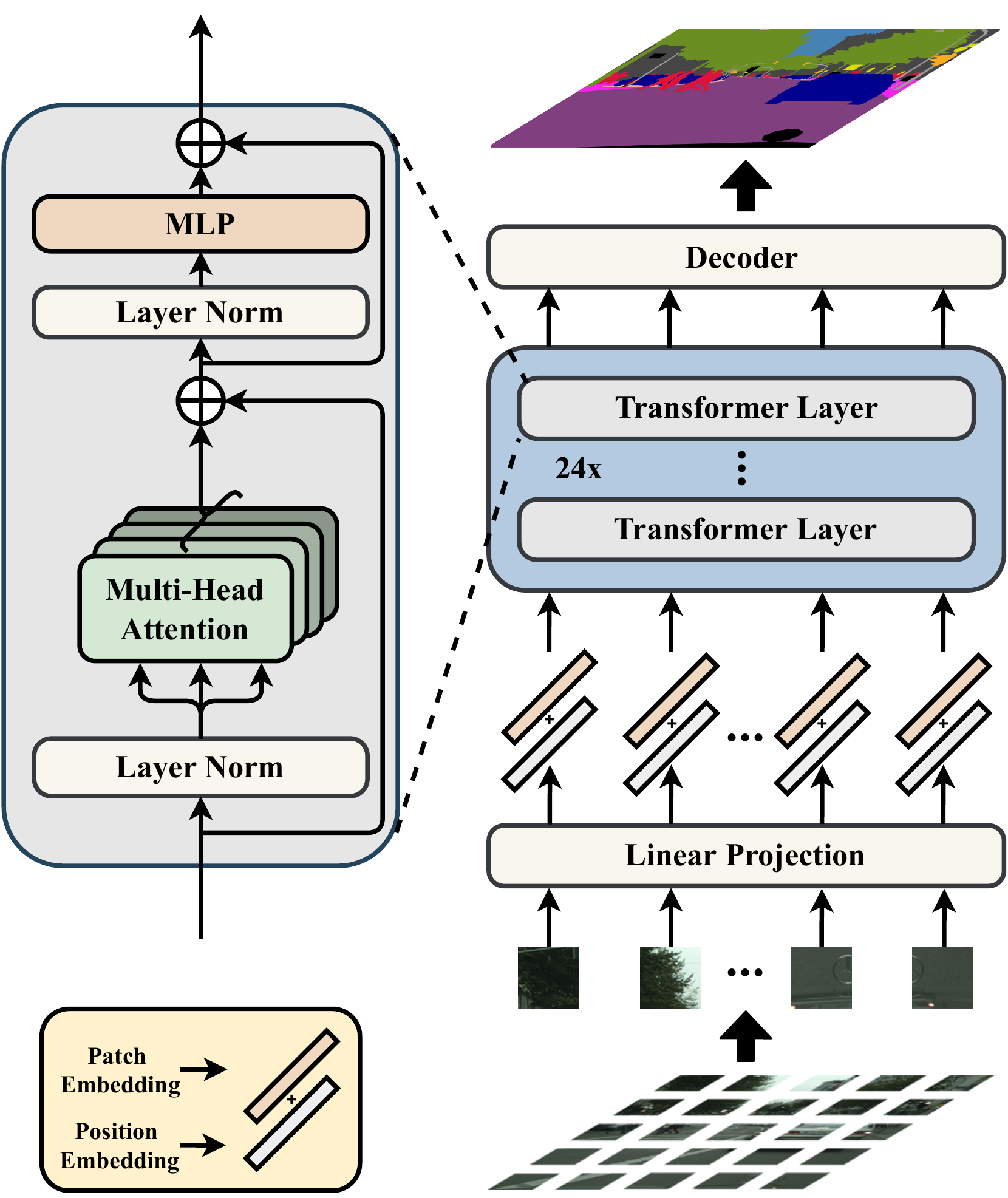}}
\end{minipage}
\medskip
\begin{minipage}[b]{.55\linewidth}
    \centering
    \subfloat[][]{\label{Genelecs:Genelec 8020 AP}\includegraphics[width=0.95\linewidth]{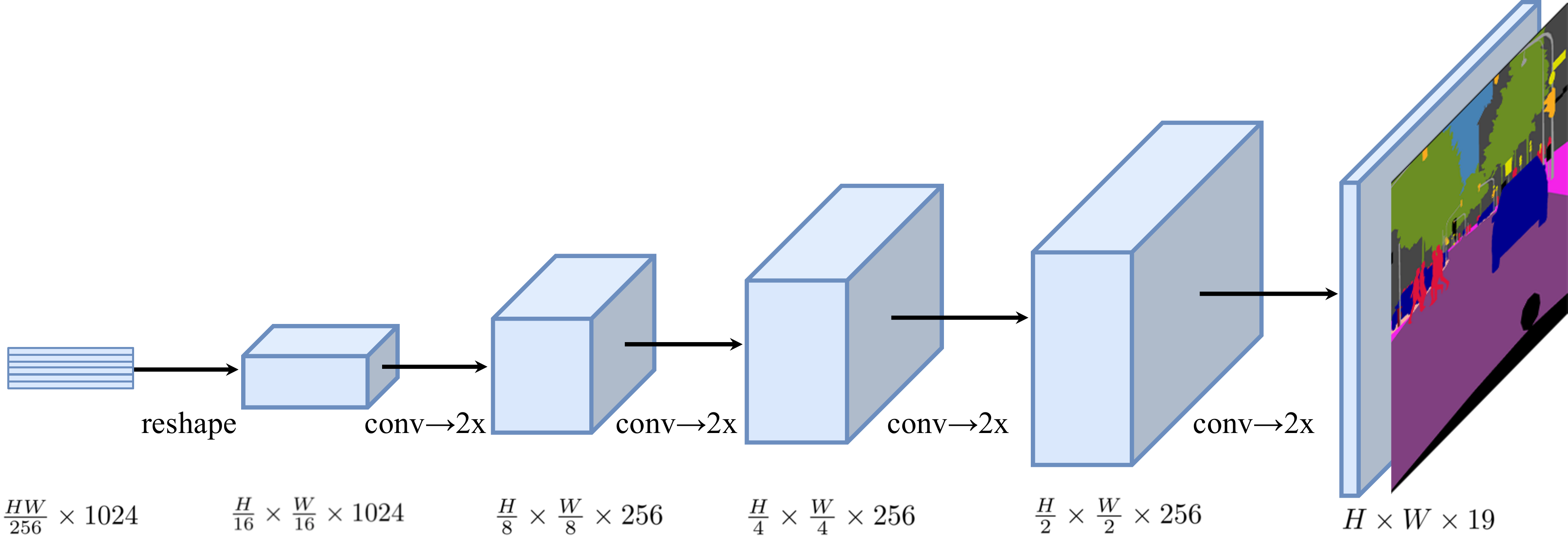}}
    
    \subfloat[][]{\label{Genelecs:Genelec 8030 AP}\includegraphics[width=0.95\linewidth]{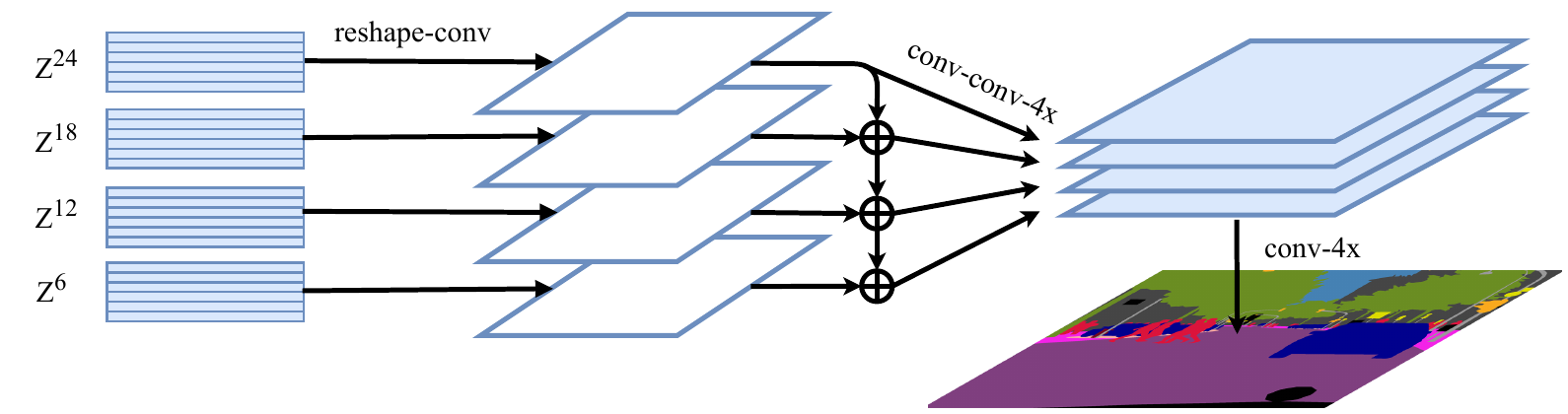}}
\end{minipage} 

\caption{\textbf{Schematic illustration of the proposed 
\modelFull~(\model)} (a).
We first split an image into fixed-size patches, linearly embed each of them, add position embeddings, and feed the resulting sequence of vectors to a standard Transformer encoder. 
To perform pixel-wise segmentation, we introduce different decoder designs:
(b) progressive upsampling (resulting in a variant called \pupModel); and
(c) multi-level feature aggregation (a variant called \mlaModel).
}
\vspace{-0.3cm}
\label{fig:SETR_transformer}
\end{figure*}

%% file: files/3-method.tex
\section{Method}

\subsection{FCN-based semantic segmentation}
In order to contrast with our new model design, let us first revisit the conventional FCN \cite{fcn} for image semantic segmentation. An FCN encoder consists of a stack of sequentially connected convolutional layers. The first layer takes as input the image, denoted as $H \times W \times 3$ with 
 $H \times W $ specifying the image size in pixels.  
The input of subsequent layer $i$ is a three-dimensional tensor sized $h \times w \times d$, where $h$ and $w$ are spatial dimensions of feature maps, and $d$ is the feature/channel dimension.
Locations of the tensor in a higher layer are computed based on the locations of tensors of all lower layers they are connected to via layer-by-layer convolutions, which are defined as their {\em receptive fields}.
Due to the locality nature of convolution operation, the receptive field increases linearly along the depth of layers, conditional on the kernel sizes (typically $3 \times 3$).
As a result, only higher layers with big receptive fields can model long-range dependencies
in this FCN architecture.
However, it is shown that the benefits of adding more layers
would diminish rapidly once reaching certain depths \cite{resnet}. Having limited receptive fields for context modeling is thus an intrinsic limitation of the vanilla FCN architecture. 

Recently, a number of state-of-the-art methods \cite{zhang2019dual,huang2018ccnet,zhang2020dynamic}
suggest that combing FCN with attention mechanism is a more effective strategy for learning long-range 
contextual information.
These methods limit the attention learning
to higher layers with smaller input sizes alone due to its quadratic complexity
\wrt the pixel number of feature tensors.
This means that dependency learning on lower-level feature tensors is lacking, leading to sub-optimal representation learning.
To overcome this limitation, we propose a pure self-attention based encoder, named {\em SEgmentation TRansformers} (SETR).

\subsection{Segmentation transformers (SETR)}
\paragraph{Image to sequence} 
\model~follows the same input-output structure as in NLP for transformation between 1D sequences. There thus exists a mismatch between 2D image and 1D sequence. 
Concretely, the Transformer, as depicted in  Figure~\ref{fig:SETR_transformer}(a), accepts a 1D sequence of feature embeddings $Z \in \mathbb{R}^{L\times C}$ as input, $L$ is the length of sequence, $C$ is the hidden channel size. Image sequentialization is thus needed to convert an input image $x\in \mathbb{R}^{H\times W\times 3}$ into $Z$.

A straightforward way for image sequentialization is 
to flatten the image pixel values into a 1D vector with size of $3HW$.
For a typical image sized at $480 (H)\times480(W)\times3$,
the resulting vector will have a length of 691,200. 
Given the quadratic model complexity of Transformer,
it is not possible that such high-dimensional vectors 
can be handled in
both space and time. 
Therefore tokenizing every single pixel as input to our transformer is out of the question.  

In view of the fact that a typical encoder designed for semantic segmentation would downsample a 2D image $x\in \mathbb{R}^{H\times W\times 3}$ into a feature map $x_f\in \mathbb{R}^{\frac{H}{16}\times \frac{W}{16}\times C}$, we  thus decide to set the transformer input sequence length $L$ as $\frac{H}{16} \times \frac{W}{16} = \frac{HW}{256}$. 
This way, the output sequence of the transformer can be simply reshaped to the target feature map $x_f$.

To obtain the $\frac{HW}{256}$-long input sequence, we divide an image $x\in \mathbb{R}^{H\times W\times 3}$ into a grid of $\frac{H}{16}\times\frac{W}{16}$ patches uniformly,
and then flatten this grid into a sequence.
By further mapping each vectorized patch $p$ into a latent $C$-dimensional embedding space
using a linear projection function $f$:
$p \longrightarrow e \in \mathbb{R}^C$, 
we obtain a 1D sequence of patch embeddings
for an image $x$.
To encode the patch spacial information,
we learn a specific embedding $p_i$ for every location $i$
which is added to $e_i$ to form the final sequence input $E = \{e_1+p_1,~e_2+p_2,~\cdots,~e_L+p_L\}$. This way, spatial information is kept despite the orderless self-attention nature of transformers.

\paragraph{Transformer}
Given the 1D embedding sequence $E$ as input,
a pure transformer based encoder is employed to learn feature representations. This means each transformer layer has a global 
 receptive field, solving the limited receptive field problem of existing FCN encoder once and for all.
The transformer encoder consists of $L_e$ layers of multi-head
self-attention (MSA) and Multilayer Perceptron (MLP) blocks \cite{velivckovic2017graph}
(Figure \ref{fig:SETR_transformer}(a)).
At each layer $l$, the input to self-attention is in a triplet of
(\texttt{query}, \texttt{key}, \texttt{value})
computed from the input $Z^{l-1} \in \mathbb{R}^{L\times C}$ as:
\begin{align}\footnotesize
    \text{query} = {Z^{l-1}} \textbf{W}_Q, \; 
    \text{key} = {Z^{l-1}} \textbf{W}_K, \;
    \text{value} = {Z^{l-1}} \textbf{W}_V,
\end{align}
where $\textbf{W}_Q$/$\textbf{W}_K$/$\textbf{W}_V  \in \mathbb{R}^{C\times d}$
are the learnable parameters of three linear projection layers
and $d$ is the dimension of (\texttt{query}, \texttt{key}, \texttt{value}).
Self-attention (SA) is then formulated as:
\begin{equation}\footnotesize
    SA(Z^{l-1}) = Z^{l-1} + 
    \operatorname{softmax}(\frac{Z^{l-1} \textbf{W}_Q (Z \textbf{W}_K)^\top}{\sqrt{d}}) (Z^{l-1} \textbf{W}_V).
    \label{eq:attn}
\end{equation}
MSA is an extension with $m$ independent SA operations
and project their concatenated outputs:
$MSA(Z^{l-1}) = [SA_1(Z^{l-1});~SA_2(Z^{l-1}); ~\cdots;~SA_m(Z^{l-1})]\textbf{W}_O$, where $\textbf{W}_O \in \mathbb{R}^{md \times C} $. $d$ is typically set to $C/m$.
The output of MSA is then transformed 
by an MLP block with residual skip as the layer output as:
\begin{equation}\footnotesize
    Z^l = MSA(Z^{l-1}) + MLP(MSA(Z^{l-1})) \in \mathbb{R}^{L \times C}.
\end{equation}
Note, layer norm is applied before MSA and MLP blocks
which is omitted for simplicity.
We denote $\{Z^1,~Z^2,~\cdots,~Z^{L_e}\}$ as the features of transformer layers.

\subsection{Decoder designs}
\label{sec:decoder}
To evaluate the effectiveness of SETR's encoder feature representations $Z$, we introduce three different decoder designs to perform pixel-level segmentation.
As the goal of the decoder is to generate the segmentation results in the original 2D image space $(H \times W)$, we need to reshape the encoder's features (that are used in the decoder),
$Z$, from a 2D shape of $\frac{HW}{256} \times C$
to a standard 3D feature map $\frac{H}{16} \times \frac{W}{16} \times C$. Next, we briefly describe the three decoders. 

\paragraph{(1) Naive upsampling (Naive)}
This naive decoder 
first projects the transformer feature $Z^{L_e}$ to the dimension of category number (\eg, 19 for experiments on Cityscapes).
For this we adopt a simple 2-layer network with architecture:
$1 \times 1$ conv + sync batch norm (w/ ReLU) + $1 \times 1$ conv.
After that, we
simply bilinearly upsample the output to the full image resolution,
followed by a classification layer with pixel-wise cross-entropy loss.
When this decoder is used,
we denote our model as \naiveModel.

\paragraph{(2) Progressive UPsampling (PUP)}
Instead of one-step upscaling which may introduce  noisy
predictions, we consider a {\em progressive upsampling} strategy 
that alternates conv layers and upsampling operations.
To maximally mitigate the adversarial effect,
we restrict upsampling to 2$\times$.
Hence, a total of 4 operations are needed for reaching 
the full resolution from $Z^{L_e}$ with size $\frac{H}{16} \times \frac{W}{16}$.
More details of this process are given in Figure \ref{fig:SETR_transformer}(b).
When using this decoder,
we denote our model as \pupModel.

\paragraph{(3) Multi-Level feature Aggregation (MLA)}

The third design is characterized by multi-level feature aggregation (Figure \ref{fig:SETR_transformer}(c))
in similar spirit of feature pyramid network \cite{fpn,kirillov2019panoptic}.
However, our decoder is  fundamentally different
because the feature representations $Z^l$ of every SETR's layer share the same resolution without a pyramid shape.

Specifically, we take as input the feature representations
$\{ Z^m\}$ ($m \in \{{\frac{L_e}{M}}, {2\frac{L_e}{M}}, \cdots, {M\frac{L_e}{M}} \}$)
from $M$ layers uniformly distributed across the layers with step $\frac{L_e}{M}$ to the decoder.
$M$ streams are then deployed, with each focusing on one specific
selected layer.
In each stream, 
we first reshape the encoder's feature
$Z^{l}$ from a 2D shape of $\frac{HW}{256} \times C$
to a 3D feature map $\frac{H}{16} \times \frac{W}{16} \times C$.
A 3-layer (kernel size $1\times 1$, $3\times 3$, and $3 \times 3$) network is applied with the feature channels 
halved at the first and third layers respectively,
and the spatial resolution upscaled $4\times$ by bilinear operation
after the third layer.
To enhance the interactions across different streams, we introduce a top-down aggregation design via element-wise addition after the first layer.
An additional $3 \times 3$ conv is applied after the element-wise additioned feature.
After the third layer, we obtain the fused feature from all the streams via channel-wise concatenation which is then bilinearly upsampled $4\times$ to the full resolution.
When using this decoder, we denote our model as \mlaModel.

%% file: files/4-experiment.tex
\section{Experiments}
\label{sec:exp}
\input{table/model_variant}
\input{table/setting}
\input{table/ablation_camera}

\subsection{Experimental setup}
We conduct experiments on three widely-used semantic segmentation benchmark datasets.

\paragraph{Cityscapes}\cite{Cityscapes}
densely annotates 19 object categories in images with urban scenes. 
It contains 5000 finely annotated images, split into 2975, ~500 and 1525 for training, validation and testing respectively. 
The images are all captured at a high resolution of $2048 \times 1024$.
In addition, it provides 19,998 coarse annotated images for model training.

\paragraph{ADE20K}\cite{ADE20K} 
is a challenging scene parsing benchmark with 150 fine-grained semantic concepts.
It contains 20210, ~2000 and 3352 images for training, validation and testing.

\paragraph{PASCAL Context}\cite{mottaghi_cvpr14} 
provides pixel-wise semantic labels for the whole scene
(both “thing” and “stuff” classes), and contains 4998 and 5105 images for training and validation respectively.
Following previous works, we evaluate on the most frequent 59 classes and the background class (60 classes in total).

\input{table/pretrain}

\paragraph{Implementation details}
Following the default setting (\eg,~data augmentation and training schedule) of public codebase {\em mmsegmentation} \cite{mmsegmentation},
(i) we apply random resize with ratio between 0.5 and 2, random cropping (768,~512 and 480 for Cityscapes, ADE20K and Pascal Context respectively) and random horizontal flipping during training for all the experiments;
(ii) We set batch size 16 and the total iteration to 160,000 and 80,000 for the experiments on ADE20K and Pascal Context.
For Cityscapes, we set batch size to 8 with a number of training schedules reported in Table~\ref{tab:ablation},~\ref{tab:city_val} and~\ref{tab:citytest} for fair comparison.
We adopt a polynomial learning rate decay schedule~\cite{pspnet} and employ SGD as the optimizer.
Momentum and weight decay are set to 0.9 and 0 respectively for all the experiments on the three datasets.
We set initial learning rate 0.001 on ADE20K and Pascal Context, and 0.01 on Cityscapes.

\paragraph{Auxiliary loss}
As~\cite{pspnet} we also find the auxiliary segmentation loss helps the model training.
Each auxiliary loss head follows a 2-layer network. 
We add auxiliary losses at different Transformer layers:
\naiveModel~($Z^{10}, Z^{15}, Z^{20}$),
\pupModel~($Z^{10}, Z^{15}, Z^{20}, Z^{24}$),
\mlaModel~($Z^6, Z^{12}, Z^{18}, Z^{24}$).
Both auxiliary loss and main loss heads are applied
concurrently.

\input{figure/visual_ade20k}
\input{table/ade20k}
\paragraph{Multi-scale test}
We use the default settings of {\em mmsegmentation} \cite{mmsegmentation}.
Specifically, the input image is first scaled to a uniform size.
Multi-scale scaling and random horizontal flip are then performed on the image with a scaling factor (0.5, 0.75, 1.0, 1.25, 1.5, 1.75). 
Sliding window is adopted for test (\eg, $480 \times 480$ for Pascal Context).
If the shorter side is smaller than the size of the sliding window, the image is scaled with its shorter side to the size of the sliding window (\eg, 480) while keeping the aspect ratio.
Synchronized BN is used in decoder and auxiliary loss heads.
For training simplicity, we do not adopt the widely-used tricks such as OHEM~\cite{ocnet} loss in model training.

\paragraph{Baselines}
We adopt dilated FCN~\cite{fcn} and Semantic FPN~\cite{kirillov2019panoptic} as baselines with their results taken from \cite{mmsegmentation}.
Our models and the baselines are trained and tested in the same settings for fair comparison.
In addition, state-of-the-art models are also compared. 
Note that the dilated FCN is with output stride 8 and we use output stride 16 in all our models due to GPU memory constrain.

\paragraph{SETR variants}

Three variants of our model with different decoder designs (see Sec.~\ref{sec:decoder}), namely \naiveModel, \pupModel~  and \mlaModel.
Besides,
we use two variants of the encoder ``T-Base'' and ``T-Large'' with 12 and 24 layers respectively (Table~\ref{tab:config}).
Unless otherwise specified, we use ``T-Large'' as the encoder for \naiveModel, \pupModel~and \mlaModel.
We denote \smallNaiveModel~as the model utilizing ``T-Base'' in \naiveModel.

Though designed as a model with a pure transformer encoder, we also set a hybrid baseline \hybridModel ~by using a ResNet-50 based FCN encoder and feeding its output feature into \model.
To cope with the GPU memory constraint and for fair comparison, we only consider `T-Base'' in \hybridModel~and set the output stride of FCN to $1/16$.
That is, \hybridModel~ is a combination of ResNet-50 and \smallNaiveModel.

\input{figure/visual_pascal_context}
\input{table/pascal-context}

\paragraph{Pre-training}
We use the pre-trained weights provided by ViT~\cite{dosovitskiy2020image} or DeiT~\cite{touvron2020training}
to initialize all the transformer layers and the input linear projection layer in our model.
We denote~\NaiveDeit~as the model utilizing DeiT~\cite{touvron2020training} pre-training in \smallNaiveModel.
All the layers without pre-training are randomly initialized.
For the FCN encoder of \hybridModel, we use the initial weights pre-trained on ImageNet-1k.
For the transformer part, we use the weights pre-trained by ViT~\cite{dosovitskiy2020image}, DeiT~\cite{touvron2020training} or randomly initialized.

\input{figure/visual_cityscapes}
\input{table/cityscape_val}

We use patch size $16\times16$ for all the experiments. We perform 2D interpolation on the pre-trained position embeddings, according to their location in the original image for different input size fine-tuning.

\paragraph{Evaluation metric}
Following the standard evaluation protocol~\cite{Cityscapes}, the metric of mean Intersection over Union (mIoU) averaged over all classes is reported.
For ADE20K, additionally pixel-wise accuracy is reported following the existing practice.

\subsection{Ablation studies}

Table~\ref{tab:ablation} and~\ref{tab:pretrain} show ablation studies on 
{\bf(a)} different variants of \model~on various training schedules,
{\bf(b)} comparison to FCN~\cite{mmsegmentation} and Semantic FPN~\cite{mmsegmentation},
{\bf(c)} pre-training on different data,
{\bf(d)} comparison with \hybridModel,
{\bf(e)} compare to FCN with different pre-training.
Unless otherwise specified, all experiments on Table~\ref{tab:ablation} and~\ref{tab:pretrain} are trained on Cityscapes train fine set with batch size 8, and evaluated using the single scale test protocol on the Cityscapes validation set in mean IoU (\%) rate.
Experiments on ADE20K also follow the single scale test protocol.
\input{table/cityscape_test}

From Table~\ref{tab:ablation}, we can make the following observations: 
{\bf(i)} Progressively upsampling the feature maps, \pupModel~achieves the best performance among all the variants on Cityscapes.
One possible reason for inferior performance of \mlaModel~is that the feature outputs of different transformer layers do not have the benefits of resolution pyramid as in feature pyramid network (FPN) (see Figure~\ref{fig:layer}).
However, \mlaModel~performs slightly better than \pupModel, and much superior to the variant \naiveModel~that upsamples the transformers output feature by 16$\times$ in one-shot, on ADE20K val set (Table~\ref{tab:pretrain} and~\ref{tab:ade}).
{\bf(ii)} The variants using ``T-Large'' (\eg,~\mlaModel~and \naiveModel) are superior to their ``T-Base'' counterparts, \ie, \smallMlaModel~and \smallNaiveModel, as expected.
{\bf(iii)}
While our \smallPupModel~(76.71) performs worse than \hybridModelB~(76.76), it shines (78.02) when training with more iterations (80k).
It suggests that FCN encoder design can be replaced in semantic segmentation, and further confirms the effectiveness of our model.
{\bf(iv)} Pre-training is critical for our model.
Randomly initialized~\pupModel~only gives 42.27\% mIoU on Cityscapes.
Model pre-trained with DeiT~\cite{touvron2020training} on ImageNet-1K gives the best performance on Cityscapes, slightly better than the counterpart pre-trained with ViT~\cite{dosovitskiy2020image} on ImageNet-21K.
{\bf(v)} 
To study the power of pre-training and further verify the effectiveness of our proposed approach, we conduct the ablation study on the pre-training strategy in Table~\ref{tab:pretrain}.
For fair comparison with the FCN baseline,
we first pre-train a ResNet-101 on the Imagenet-21k dataset with a classification task and then adopt the pre-trained weights for a dilated FCN training for the semantic segmentation task on ADE20K or Cityscapes.
Table~\ref{tab:pretrain} shows that with ImageNet-21k pre-training FCN baseline experienced a clear improvement over the variant pre-trained on ImageNet-1k.
However, our method outperforms the FCN counterparts by a large margin, verifying that the advantage of our approach largely comes from the proposed \textit{sequence-to-sequence} modeling strategy rather than bigger pre-training data.

\input{figure/visual_layer}

\subsection{Comparison to state-of-the-art }

\paragraph{Results on ADE20K}
Table~\ref{tab:ade} presents our results on the more challenging ADE20K dataset.
Our \mlaModel~achieves superior mIoU of 48.64\% with single-scale (SS) inference.
When multi-scale inference is adopted, our method achieves a new state of the art with mIoU hitting 50.28\%.
Figure~\ref{fig:ade} shows the qualitative results of our model and dilated FCN on ADE20K.
When training a single model on the train+validation set with the default 160,000 iterations, our method ranks $1^{st}$
place in the highly competitive ADE20K test server leaderboard.

\input{figure/visual_attention}

\paragraph{Results on Pascal Context}
Table~\ref{tab:pascal} compares the segmentation results on Pascal Context.
Dilated FCN with the ResNet-101 backbone achieves a mIoU of 45.74\%.
Using the same training schedule, our proposed SETR significantly outperforms this baseline, achieving mIoU of 54.40\% (\pupModel) and 54.87\% (\mlaModel).
\mlaModel~further improves the performance to 55.83\% when multi-scale (MS) inference is adopted, outperforming the nearest rival APCNet with a clear margin.
Figure~\ref{fig:pascal} gives some qualitative results
of SETR and dilated FCN. 
Further visualization of the learned attention maps in Figure~\ref{fig:attention} shows that \model~can attend to semantically meaningful foreground regions, demonstrating its ability to learn discriminative feature representations useful for segmentation.

\paragraph{Results on Cityscapes}
Tables~\ref{tab:city_val} and~\ref{tab:citytest} show the comparative results on the validation and test set of  Cityscapes respectively. 
We can see that our model \pupModel~is superior to FCN baselines, and FCN plus attention based approaches, such as Non-local~\cite{wang2018nonlocal} and CCNet~\cite{huang2018ccnet}; and its performance is on par with the best results reported so far. 
On this dataset we can now compare with the closely related Axial-DeepLab~\cite{cheng2020panoptic,wang2020axial} which aims to use an attention-alone model but still follows the basic structure of FCN. 
Note that Axial-DeepLab sets the same output stride 16 as ours.
However, its full input  resolution ($1024 \times 2048$) is much larger than our crop size $768 \times 768$, and it runs more epochs (60k iteration with batch size 32) than our setting (80k iterations with batch size 8). 
Nevertheless, our model is still superior to Axial-DeepLab 
when multi-scale inference is adopted on Cityscapes validation set.
Using the fine set only, our model (trained with 100k iterations) outperforms Axial-DeepLab-XL with a clear margin on the test set.
Figure~\ref{fig:city} shows the qualitative results of our model and dilated FCN on Cityscapes.

%% file: table/model_variant.tex
\begin{table}[t]
\centering
    \begin{tabular}[t]{c|ccc}
        Model &  T-layers & Hidden size & Att head \\
        \shline
        T-Base &12 & 768 & 12 \\
        T-Large &24 & 1024 & 16 \\
    \end{tabular}

\caption{Configuration of Transformer backbone variants.}
\label{tab:config}
\end{table}

%% file: table/setting.tex
\definecolor{Gray}{gray}{0.5}
\newcommand{\randinit}{\tablestyle{1pt}{1} \begin{tabular}{z{21}y{26}} \multicolumn{2}{c}{\demph{random init.}} \end{tabular}}
\newcommand{\mocoimgnet}{\tablestyle{0pt}{1} \begin{tabular}{z{21}y{26}} \textbf{MoCo} & ~~IN-1M \end{tabular}}
\newcommand{\mocoins}{\tablestyle{0pt}{1} \begin{tabular}{z{21}y{26}} \textbf{MoCo} & ~~IG-1B \end{tabular}}
\newcommand{\supimgnet}{\tablestyle{0pt}{1} \begin{tabular}{z{21}y{26}} super. & ~~IN-1M \end{tabular}}

\newcommand{\demph}[1]{\textcolor{Gray}{#1}}
\newcommand{\std}[1]{{\fontsize{5pt}{1em}\selectfont ~~$_\pm$$_{\text{#1}}$}}

\definecolor{Highlight}{HTML}{39b54a}  

\renewcommand{\hl}[1]{\textcolor{Highlight}{#1}}

\newcommand{\res}[3]{
\tablestyle{1pt}{1}
\begin{tabular}{z{25}y{30}}
{#1} &
\fontsize{7.5pt}{1em}\selectfont{~(${#2}${#3})}
\end{tabular}}

\newcommand{\reshl}[3]{
\tablestyle{1pt}{1} 
\begin{tabular}{z{25}y{30}}
{#1} &
\fontsize{7.5pt}{1em}\selectfont{~\hl{(${#2}$\textbf{#3})}}
\end{tabular}}

\newcommand{\resrand}[2]{\tablestyle{1pt}{1} \begin{tabular}{z{25}y{30}} \demph{#1} & {} \end{tabular}}
\newcommand{\ressup}[2]{\tablestyle{1pt}{1} \begin{tabular}{z{25}y{30}} {#1} & {} \end{tabular}}

%% file: table/ablation_camera.tex
\begin{table}[t]
\centering
\small
\setlength{\tabcolsep}{0.4em}
\begin{tabular}{lc|c|c|cc}
\multicolumn{1}{c}{Method} & \multicolumn{1}{c|}{Pre} & Backbone   & \#Params & 40k   & 80k   \\
\shline
FCN~\cite{mmsegmentation}                        & 1K                   & R-101 &   68.59M     &   73.93    &  75.52     \\
Semantic FPN~\cite{mmsegmentation}               & 1K                  & R-101 &  47.51M        &   -  &   75.80    \\
\hline
\hybridModelB               & R                  & T-Base    &   112.59M &74.48  &77.36  \\
\hybridModelB               & 21K                  & T-Base    &   112.59M &76.76  &76.57  \\
\hybridModelD               & 21K                  & T-Base    &   112.59M &77.42  &78.28  \\
\hline
\naiveModel                & 21K                  & T-Large    & 305.67M   & 77.37 & 77.90 \\
\mlaModel                   & 21K                  & T-Large    & 310.57M   & 76.65 & 77.24 \\
\pupModel                   & 21K                  & T-Large    & 318.31M   & 78.39 & 79.34 \\
\hline
\pupModel                & R                  & T-Large    & 318.31M   & 42.27 & - \\
\hline
\smallNaiveModel           & 21K                  & T-Base     & 87.69M    & 75.54 & 76.25 \\
\smallMlaModel                 & 21K                  & T-Base     & 92.59M    & 75.60 & 76.87 \\
\smallPupModel                 & 21K                  & T-Base     & 97.64M    & 76.71 & 78.02 \\
\hline
\NaiveDeit            & 1K                   & T-Base     & 87.69M    & 77.85      &   78.66    \\
\MlaDeit              & 1K                   & T-Base     & 92.59M    & 78.04      &    78.98   \\
\PupDeit              & 1K                   & T-Base     & 97.64M    &  {\bf 78.79}     &    {\bf 79.45}  
\end{tabular}
\caption{\textbf{Comparing~\model~variants} on different pre-training strategies and backbones.
All experiments are trained on Cityscapes train fine set with batch size 8, and evaluated using the single scale test protocol on the Cityscapes validation set in mean IoU (\%) rate.
``Pre'' denotes the pre-training of transformer part.
``R'' means the transformer part is randomly initialized.}
\label{tab:ablation}
\end{table}

%% file: table/pretrain.tex
\begin{table}[t]
\centering
\setlength{\tabcolsep}{0.2em}
\begin{tabular}[t]{ccc|cc}
Method & Pre & Backbone & ADE20K & Cityscapes \\
\shline
\multicolumn{1}{l}{FCN ~\cite{mmsegmentation}} & 1K & R-101 & 39.91 &73.93 \\
\multicolumn{1}{l}{FCN } & 21K & R-101 & 42.17 & 76.38 \\
\hline
\multicolumn{1}{l}{\mlaModel } & 21K & T-Large & \textbf{48.64} & 76.65  \\
\multicolumn{1}{l}{\pupModel } & 21K & T-Large & 48.58 & 78.39 \\
\hline
\multicolumn{1}{l}{\MlaDeit } & 1K & T-Large &46.15  &78.98  \\
\multicolumn{1}{l}{\PupDeit } & 1K & T-Large &46.24  &\textbf{79.45}  \\
\end{tabular}
\caption{\textbf{
Comparison to FCN with different pre-training}
with single-scale inference on the ADE20K val and Cityscapes val set.}
\label{tab:pretrain}
\end{table}

%% file: figure/visual_ade20k.tex
\begin{figure}[t]\centering
\includegraphics[width=1.0\linewidth]{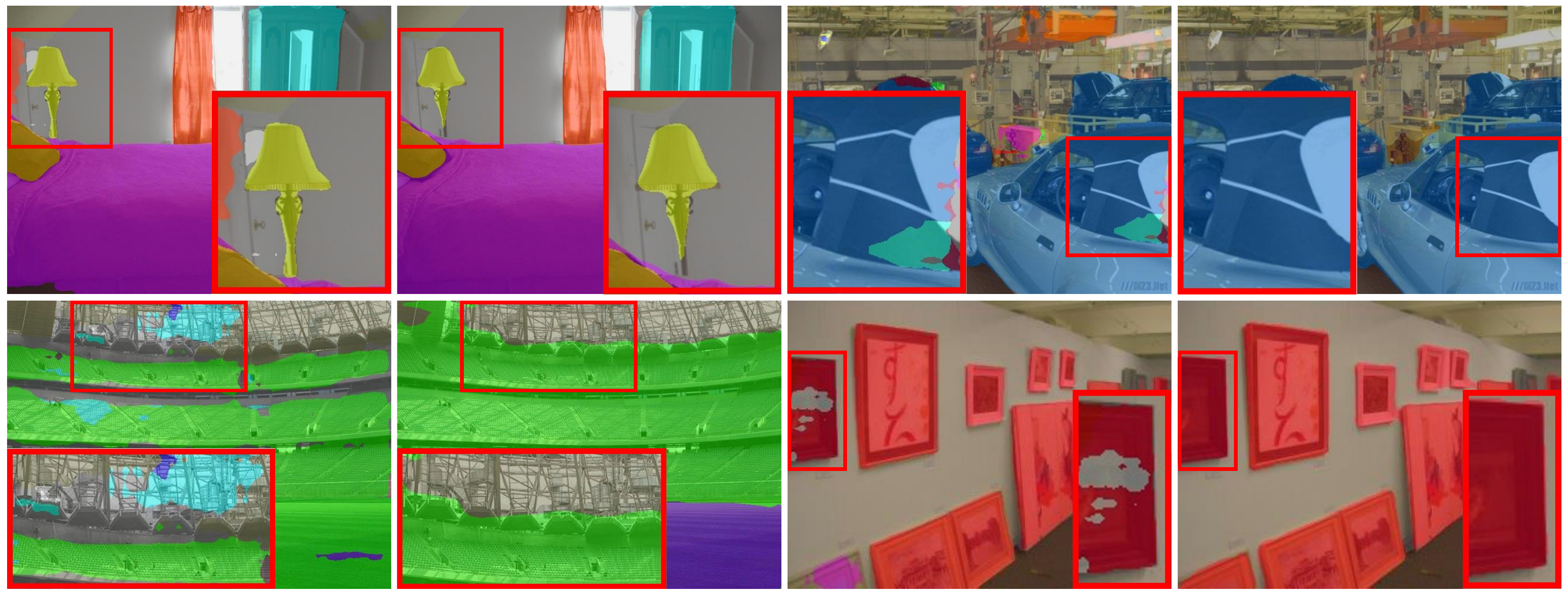}
\caption{\textbf{Qualitative results on ADE20K:} SETR (right column) vs. dilated FCN baseline (left column) in each pair. 
Best viewed in color and zoom in.}
\vspace{-0.3em}
\label{fig:ade}
\end{figure}

%% file: table/ade20k.tex
\begin{table}[t]
\tablestyle{1.8pt}{1.05}
\begin{tabular}{x{44}x{14}x{44}|x{34}|x{34}c}
Method &Pre & Backbone & \#Params & mIoU \\
\shline
\multicolumn{1}{l}{FCN (160k, SS)~\cite{mmsegmentation}}&1K & ResNet-101 &68.59M  & 39.91 \\
\multicolumn{1}{l}{FCN (160k, MS)~\cite{mmsegmentation}}&1K & ResNet-101 &68.59M  & 41.40 \\
\hline
\multicolumn{1}{l}{CCNet~\cite{huang2018ccnet}}&1K & ResNet-101 &-  & 45.22 \\
\multicolumn{1}{l}{Strip pooling~\cite{hou2020strip}}&1K & ResNet-101 &-  & 45.60 \\
\multicolumn{1}{l}{DANet~\cite{DAnet}}&1K & ResNet-101 &69.0M  &45.30  \\
\multicolumn{1}{l}{OCRNet~\cite{yuan2019object}}&1K & ResNet-101 &71.0M  &45.70  \\
\multicolumn{1}{l}{UperNet~\cite{xiao2018unified}}&1K & ResNet-101 &86.0M  &44.90  \\
\multicolumn{1}{l}{Deeplab V3+~\cite{deeplabv3p}}&1K & ResNet-101 &63.0M  &46.40 \\
\hline
\multicolumn{1}{l}{\naiveModel~(160k, SS)}&21K & T-Large &305.67M  & 48.06 \\
\multicolumn{1}{l}{\naiveModel~(160k, MS)}&21K & T-Large &305.67M  & 48.80\\
\multicolumn{1}{l}{\pupModel~(160k, SS)}&21K & T-Large&318.31M & 48.58  \\
\multicolumn{1}{l}{\pupModel~(160k, MS)}&21K & T-Large &318.31M & 50.09 \\
\multicolumn{1}{l}{\mlaModel~(160k, SS)}&21K & T-Large &310.57M & 48.64  \\
\multicolumn{1}{l}{\mlaModel~(160k, MS)}&21K & T-Large &310.57M & \textbf{50.28} \\
\multicolumn{1}{l}{\PupDeit ~(160k, SS)}&1K & T-Base &97.64M    & 46.34  \\
\multicolumn{1}{l}{\PupDeit ~(160k, MS)}&1K & T-Base &97.64M &47.30      \\
\multicolumn{1}{l}{\MlaDeit ~(160k, SS)}&1K & T-Base &92.59M &46.15     \\
\multicolumn{1}{l}{\MlaDeit ~(160k, MS)}&1K & T-Base &92.59M &47.71     \\
\end{tabular}
\caption{\textbf{State-of-the-art comparison on the ADE20K dataset.}
Performances of different model variants
are reported.
SS: Single-scale inference. MS: Multi-scale inference.}
\label{tab:ade}
\end{table}

%% file: figure/visual_pascal_context.tex
\begin{figure}[t]\centering
\includegraphics[width=1.0\linewidth]{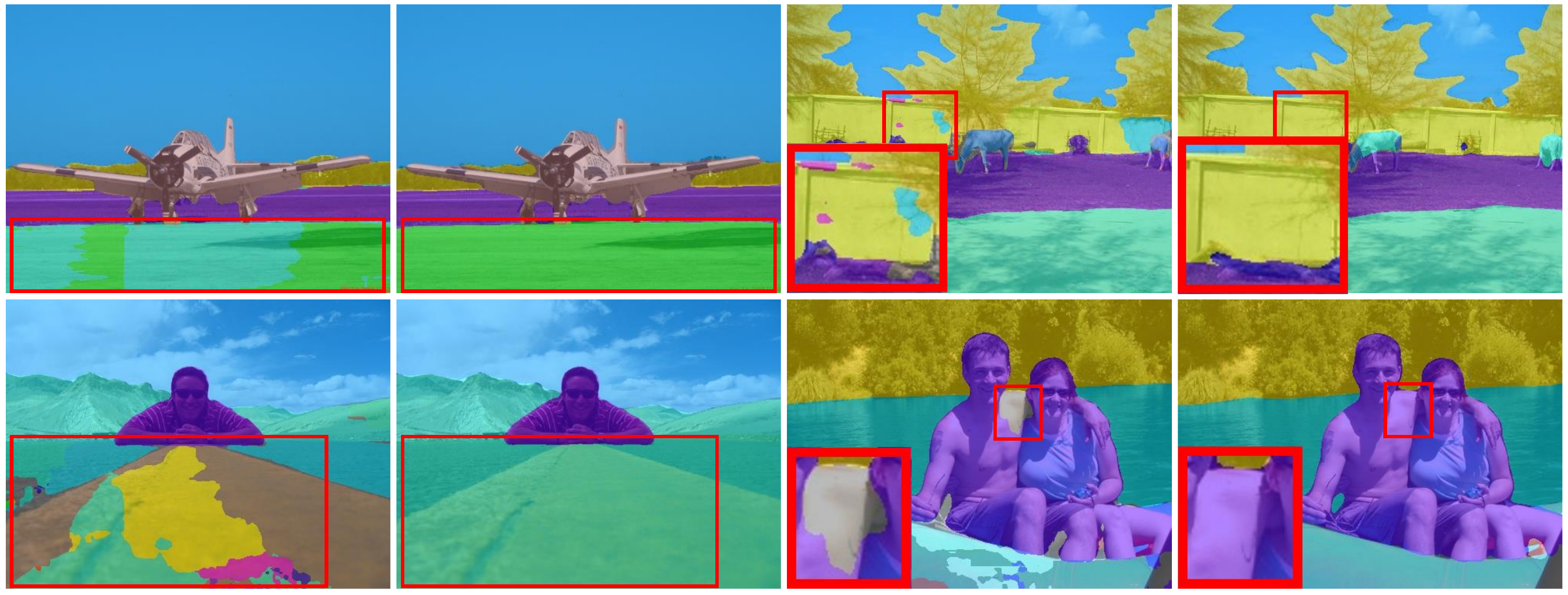}
\caption{\textbf{Qualitative results on Pascal Context:} SETR (right column) vs. dilated FCN baseline (left column) in each pair. 
Best viewed in color and zoom in.}
\vspace{-0.3em}
\label{fig:pascal}
\end{figure}

%% file: table/pascal-context.tex
\begin{table}[t]
\tablestyle{1.8pt}{1.05}
\begin{tabular}{x{64}x{34}x{54}|x{45}c|}
Method & Pre & Backbone & mIoU \\
\shline
\multicolumn{1}{l}{FCN (80k, SS)~\cite{mmsegmentation}}&1K & ResNet-101 & 44.47 \\
\multicolumn{1}{l}{FCN (80k, MS)~\cite{mmsegmentation}}&1K & ResNet-101 & 45.74 \\
\hline
\multicolumn{1}{l}{DANet~\cite{DAnet}}&1K & ResNet-101 & 52.60 \\
\multicolumn{1}{l}{EMANet~\cite{li2019expectation}}&1K & ResNet-101 & 53.10 \\
\multicolumn{1}{l}{SVCNet~\cite{ding2019semantic}}&1K & ResNet-101 & 53.20 \\
\multicolumn{1}{l}{Strip pooling~\cite{hou2020strip}}&1K & ResNet-101 & 54.50 \\
\multicolumn{1}{l}{GFFNet~\cite{gff}}&1K & ResNet-101 & 54.20 \\
\multicolumn{1}{l}{APCNet~\cite{he2019adaptive}}&1K & ResNet-101 & 54.70 \\
\hline
\multicolumn{1}{l}{\naiveModel~(80k, SS)}&21K & T-Large & 52.89 \\
\multicolumn{1}{l}{\naiveModel~(80k, MS)}&21K & T-Large & 53.61 \\
\multicolumn{1}{l}{\pupModel~(80k, SS)}&21K & T-Large & 54.40 \\
\multicolumn{1}{l}{\pupModel~(80k, MS)}&21K & T-Large & 55.27 \\
\multicolumn{1}{l}{\mlaModel~(80k, SS)}&21K & T-Large & 54.87 \\
\multicolumn{1}{l}{\mlaModel~(80k, MS)}&21K & T-Large & \textbf{55.83} \\
\multicolumn{1}{l}{\PupDeit~(80k, SS)}&1K & T-Base &  52.71\\
\multicolumn{1}{l}{\PupDeit~(80k, MS)}&1K & T-Base &  53.71\\
\multicolumn{1}{l}{\MlaDeit~(80k, SS)}&1K & T-Base &  52.91\\
\multicolumn{1}{l}{\MlaDeit~(80k, MS)}&1K & T-Base &  53.74\\
\end{tabular}
\caption{\textbf{State-of-the-art comparison on the Pascal Context dataset.}
Performances of different model variants
are reported.
SS: Single-scale inference. MS: Multi-scale inference.}
\label{tab:pascal}
\end{table}

%% file: figure/visual_cityscapes.tex
\begin{figure*}[t]\centering
\includegraphics[width=1.0\linewidth]{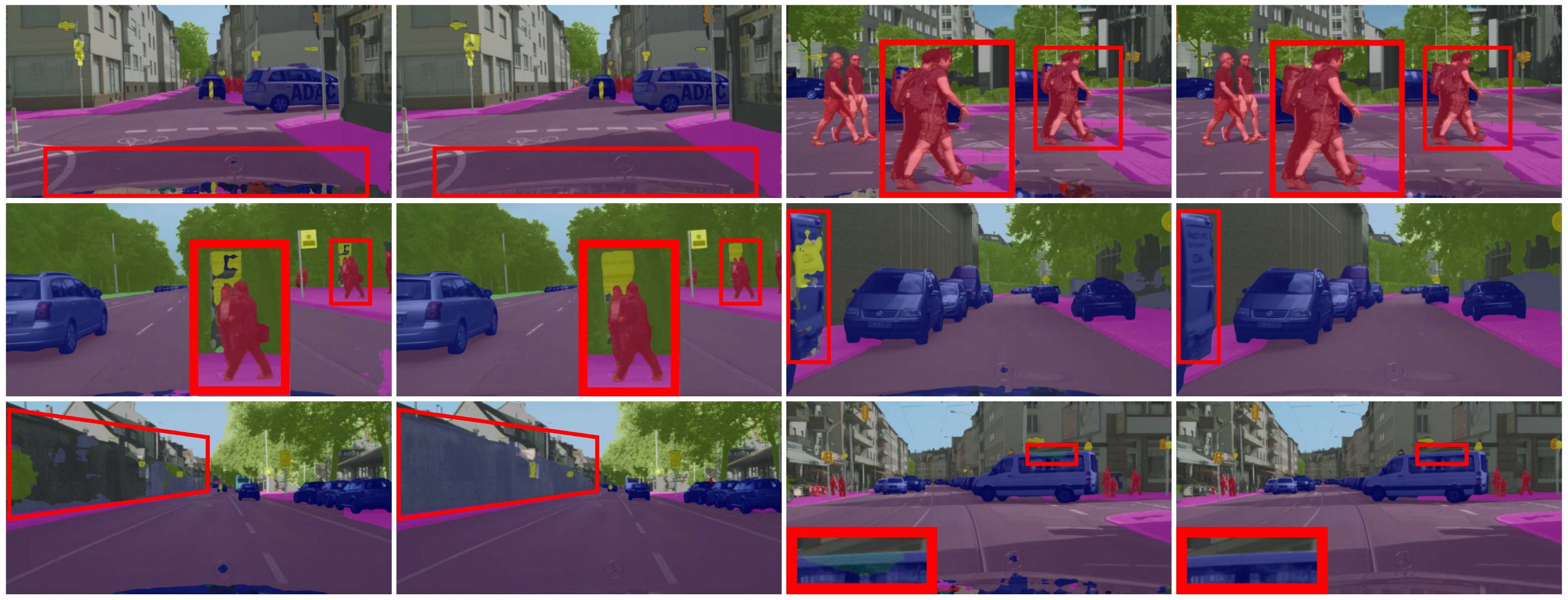}
\caption{\textbf{Qualitative results on Cityscapes:} SETR (right column) vs. dilated FCN baseline (left column) in each pair. 
Best viewed in color and zoom in.}
\vspace{-0.3cm}
\label{fig:city}
\end{figure*}

%% file: table/cityscape_val.tex
\begin{table}[t]
\tablestyle{1.8pt}{1.05}
\begin{tabular}{x{64}x{64}|x{45}c}
Method & Backbone & mIoU\\
\shline
\multicolumn{1}{l}{FCN (40k, SS)~\cite{mmsegmentation}} & ResNet-101 & 73.93 \\
\multicolumn{1}{l}{FCN (40k, MS)~\cite{mmsegmentation}} & ResNet-101 & 75.14 \\
\multicolumn{1}{l}{FCN (80k, SS)~\cite{mmsegmentation}} & ResNet-101 & 75.52 \\
\multicolumn{1}{l}{FCN (80k, MS)~\cite{mmsegmentation}} & ResNet-101 & 76.61 \\
\hline
\multicolumn{1}{l}{PSPNet~\cite{pspnet}} & ResNet-101 & 78.50 \\
\multicolumn{1}{l}{DeepLab-v3~\cite{deeplabv3} (MS)} & ResNet-101 & 79.30 \\
\multicolumn{1}{l}{NonLocal~\cite{wang2018nonlocal}} & ResNet-101 & 79.10 \\
\multicolumn{1}{l}{CCNet~\cite{huang2018ccnet}} & ResNet-101 & 80.20 \\
\multicolumn{1}{l}{GCNet~\cite{cao2019gcnet}} & ResNet-101 & 78.10 \\
\multicolumn{1}{l}{Axial-DeepLab-XL~\cite{wang2020axial} (MS)} & Axial-ResNet-XL & 81.10 \\
\multicolumn{1}{l}{Axial-DeepLab-L~\cite{wang2020axial} (MS)} & Axial-ResNet-L & 81.50 \\
\hline
\multicolumn{1}{l}{\pupModel~(40k, SS)} & T-Large & 78.39 \\
\multicolumn{1}{l}{\pupModel~(40k, MS)} & T-Large &81.57  \\
\multicolumn{1}{l}{\pupModel~(80k, SS)} & T-Large & 79.34 \\
\multicolumn{1}{l}{\pupModel~(80k, MS)} & T-Large &\textbf{82.15}  \\
\end{tabular}
\caption{\textbf{State-of-the-art comparison on the Cityscapes validation set.}
Performances of different training schedules (\eg, 40k and 80k) are reported.
SS: Single-scale inference. MS: Multi-scale inference.}
\label{tab:city_val}
\end{table}

%% file: table/cityscape_test.tex
\begin{table}[t]
\tablestyle{1.8pt}{1.05}
\begin{tabular}{x{64}x{64}|x{45}c}
Method & Backbone & mIoU\\
\shline
\multicolumn{1}{l}{PSPNet~\cite{pspnet}} & ResNet-101 & 78.40 \\
\multicolumn{1}{l}{DenseASPP~\cite{denseaspp}} & DenseNet-161 & 80.60 \\
\multicolumn{1}{l}{BiSeNet~\cite{bisenet}} & ResNet-101 & 78.90 \\
\multicolumn{1}{l}{PSANet~\cite{psanet}} & ResNet-101 & 80.10 \\
\multicolumn{1}{l}{DANet~\cite{DAnet}} & ResNet-101 & 81.50 \\
\multicolumn{1}{l}{OCNet~\cite{ocnet}} & ResNet-101 & 80.10 \\
\multicolumn{1}{l}{CCNet~\cite{huang2018ccnet}} & ResNet-101 & 81.90 \\
\multicolumn{1}{l}{Axial-DeepLab-L~\cite{wang2020axial}} & Axial-ResNet-L & 79.50 \\
\multicolumn{1}{l}{Axial-DeepLab-XL~\cite{wang2020axial}} & Axial-ResNet-XL & 79.90 \\
\hline
\multicolumn{1}{l}{\pupModel~(100k)} & T-Large & 81.08\\
\multicolumn{1}{l}{\pupModel$^\ddag$} & T-Large & 81.64 \\
\end{tabular}
\caption{\textbf{Comparison on the Cityscapes test set.}
$\ddag$: trained on fine and coarse annotated data.
}
\label{tab:citytest}
\end{table}

%% file: figure/visual_layer.tex
\begin{figure}[t]\centering
\includegraphics[width=1.02\linewidth]{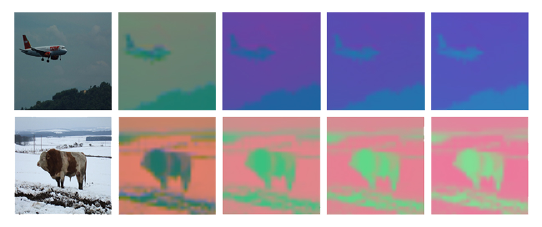}
\vspace{-0.3cm}
\caption{Visualization of output feature of layer $Z^1$, $Z^9$, $Z^{17}$, $Z^{24}$ of SETR trained on Pascal Context.
Best viewed in color.}
\label{fig:layer}
\end{figure}

%% file: figure/visual_attention.tex
\begin{figure}[t]\centering
\includegraphics[width=1.02\linewidth]{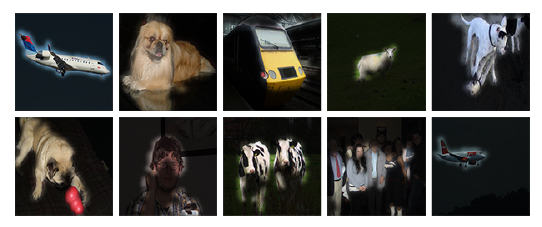}
\vspace{-0.3cm}
\caption{
Examples of attention maps from SETR trained on Pascal Context. 
}
\label{fig:attention}
\end{figure}

%% file: files/5-conclusion.tex
\section{Conclusion}
In this work, we have presented an alternative 
perspective for semantic segmentation by introducing a {\em sequence-to-sequence} prediction framework.
In contrast to existing FCN based methods that enlarge the receptive field typically with dilated convolutions and attention modules at the {\em component} level, 
we made a step change at the {\em architectural} level to completely eliminate the reliance on FCN and elegantly solve the limited receptive field challenge.
We implemented the proposed idea with Transformers
that can model global context at every stage of feature learning.
Along with a set of decoder designs in different complexity, strong segmentation models are established
with none of the bells and whistles deployed by recent methods.
Extensive experiments demonstrate that our models set new state of the art on ADE20, Pascal Context and competitive results on Cityscapes.
Encouragingly, our method is ranked the $1^{st}$ place in the highly competitive ADE20K test server leaderboard on the day of submission.

\section*{Acknowledgments}
This work was supported by Shanghai Municipal Science and Technology Major Project (No.2018SHZDZX01), 
ZJLab, 
and Shanghai Center for Brain Science and Brain-Inspired Technology.

%% file: files/6-appendix.tex
\appendix
\section{Visualizations}

\paragraph{Position embedding}
Visualization of the learned position embedding in Figure~\ref{fig:pos} shows that the model learns to encode distance within the image in the similarity of position embeddings.

\paragraph{Features}
 
Figure~\ref{fig:layer-pup} shows the feature visualization of our \pupModel.
For the encoder, 24 output features from the 24 transformer layers namely $Z^1-Z^{24}$ are collected. Meanwhile, 5 features ($U^1-U^5$) right after each bilinear interpolation in the decoder head are visited.

\paragraph{Attention maps}

Attention maps (Figure~\ref{fig:attention-50}) in each transformer layer catch our interest. 
There are 16 heads and 24 layers in T-large.
Similar to \cite{abnar2020quantifying}, a recursion perspective into this problem is applied. 
Figure~\ref{fig:ori-gt-att} shows the attention maps of different selected spatial points (red).

\input{figure/visual_position_embed_80}
\input{figure/ori-gt-att}

\input{figure/visual_layer-pup}
\input{figure/visual_attention_50}

%% file: figure/visual_position_embed_80.tex
\begin{figure}[h!]\centering
\includegraphics[width=0.92\linewidth]{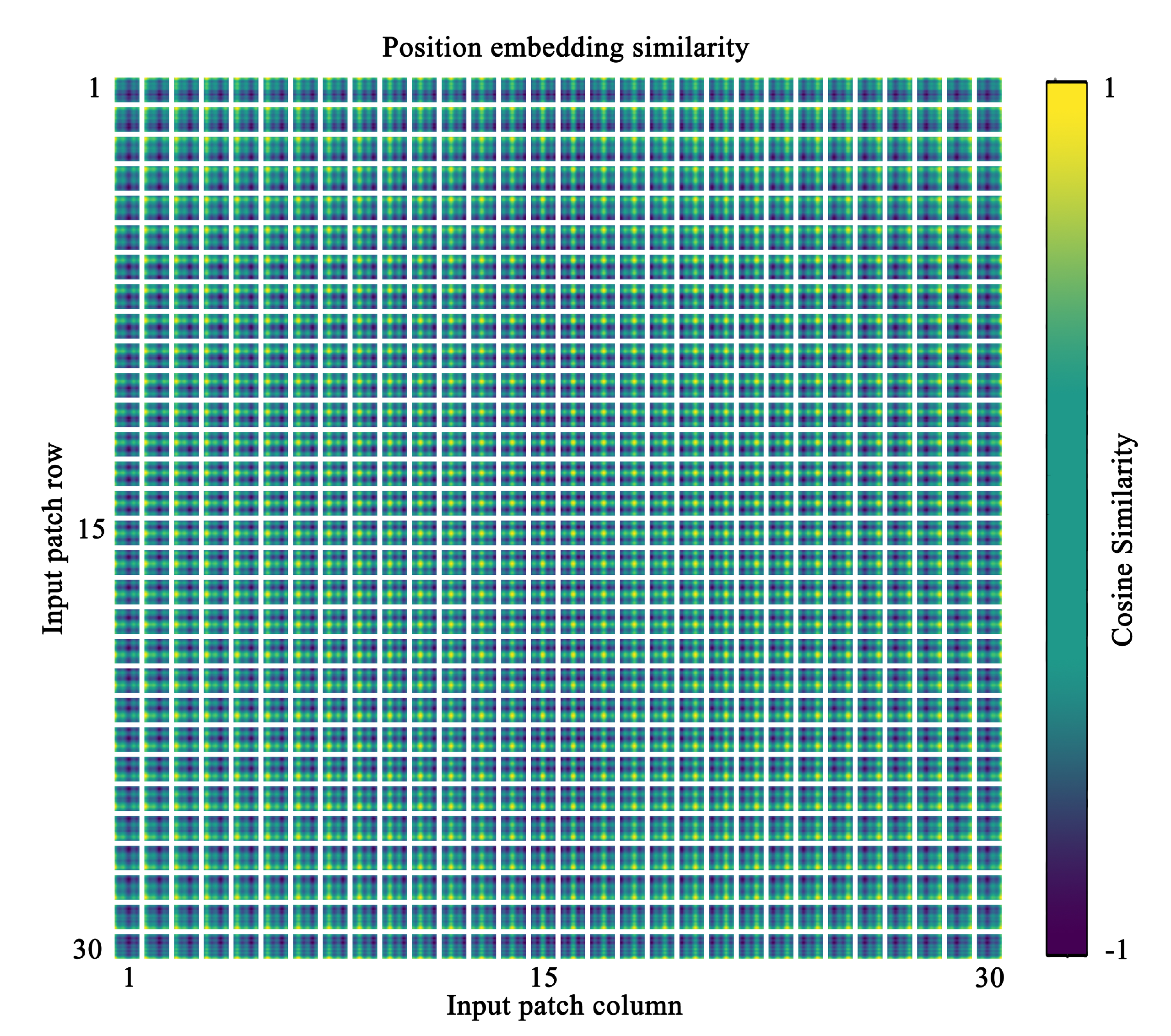}
\caption{Similarity of position embeddings of~\pupModel~trained on Pascal Context.
Tiles show the cosine similarity between the position embedding of the patch with the indicated row and column and the position embeddings of all other patches.}
\label{fig:pos}
\end{figure}

%% file: figure/ori-gt-att.tex
\begin{figure}[b]\centering
\includegraphics[width=0.75\linewidth]{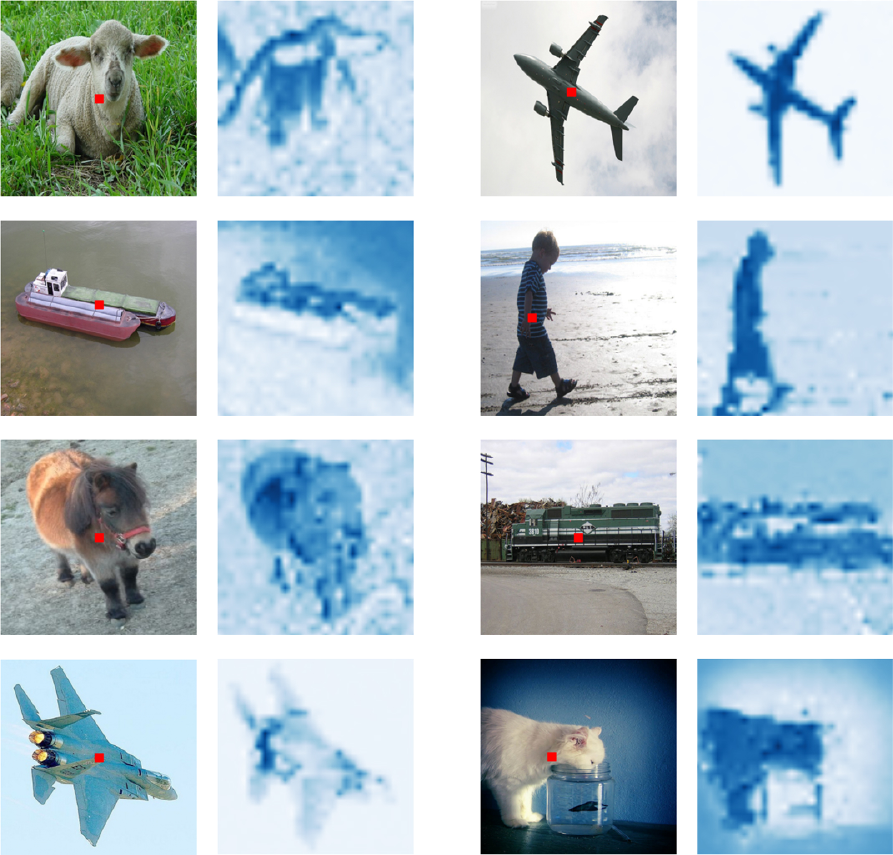}
\caption{The first and third columns show images from Pascal Context.
The second and fourth columns illustrate the attention map of the picked points (red).}
\label{fig:ori-gt-att}
\end{figure}

%% file: figure/visual_layer-pup.tex
\begin{figure*}[h!]\centering
\includegraphics[width=0.85\linewidth]{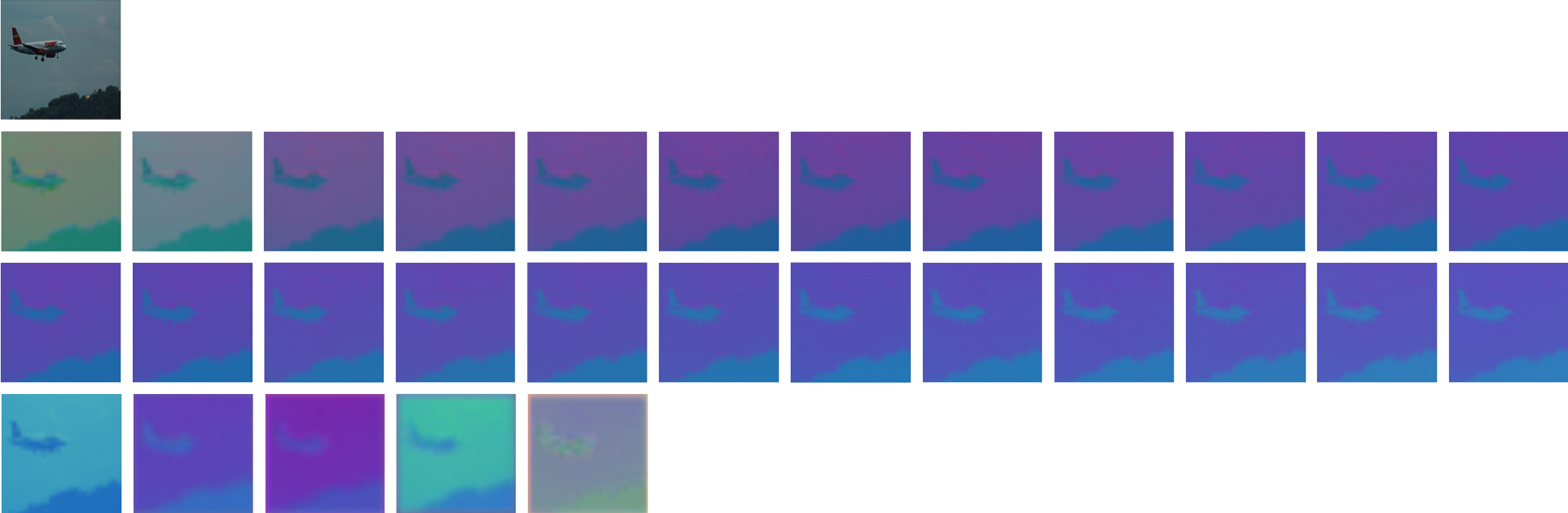}
\caption{Visualization of output feature of layer $Z^1-Z^{24}$ and $U^1-U^5$ of \pupModel~trained on Pascal Context.
Best view in color.
\textbf{First row:} The input image. 
\textbf{Second row:} Layer $Z^1$-$Z^{12}$.
\textbf{Third row:} Layer $Z^{13}$-$Z^{24}$.
\textbf{Fourth row:} Layer $U^1-U^5$.
} 
\label{fig:layer-pup}
\end{figure*}

%% file: figure/visual_attention_50.tex
\begin{figure*}[h!]\centering
\includegraphics[width=0.85\linewidth]{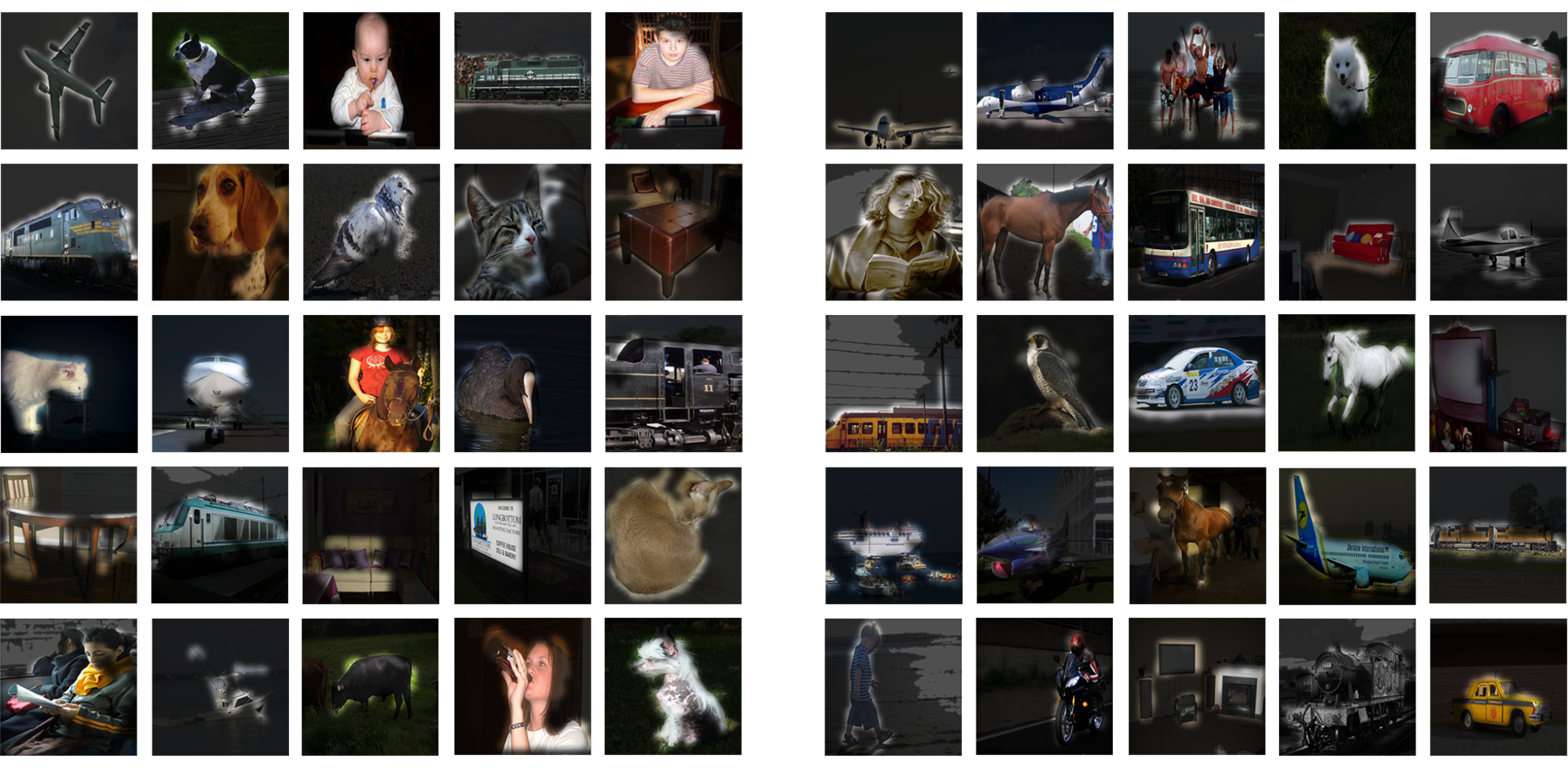}
\caption{More examples of attention maps from~\pupModel~trained on Pascal Context.}
\label{fig:attention-50}
\end{figure*}